\documentclass[letterpaper, 10 pt, conference]{ieeeconf}  % Comment this line out if you need a4paper

\IEEEoverridecommandlockouts                              % This command is only needed if 
\usepackage{graphicx} % for pdf, bitmapped graphics files
\usepackage{times} % assumes new font selection scheme installed
\usepackage{hyperref}
\usepackage{cite} 
\usepackage{amsmath}
\usepackage{multirow}
\usepackage{booktabs}
\usepackage{makecell}
\usepackage{array}
\usepackage{amssymb}

\title{\LARGE \bf
AssemMate: Graph-Based LLM for Robotic Assembly Assistance
\thanks{Video available at \url{https://youtu.be/eX5uBRrdV6s}.}}

\author{Qi Zheng\dag, Chaoran Zhang\dag, Zijian Liang\dag, EnTe Lin, Shubo Cui, Qinghongbing Xie, Zhaobo Xu, Long Zeng*% <-this % stops a space
    \thanks{\dag \ Equal contribution. (e-mail: zheng-q25@mails.tsinghua.edu.cn)}
    \thanks{* corresponding author.}
    \thanks{Qi Zheng, Chaoran Zhang, Zijian Liang, EnTe Lin, Shubo Cui, Qinghongbing Xie, Zhaobo Xu, and Long Zeng are with Tsinghua Shenzhen International Graduate School, Tsinghua University, Shenzhen, China.}
}

\begin{document}

\maketitle
\thispagestyle{empty}
\pagestyle{empty}

%%%%%%%%%%%%%%%%%%%%%%%%%%%%%%%%%%%%%%%%%%%%%%%%%%%%%%%%%%%%%%%%%%%%%%%%%%%%%%%%

\begin{abstract}

Large Language Model (LLM)-based robotic assembly assistance has gained significant research attention. It requires the injection of domain-specific knowledge to guide the assembly process through natural language interaction with humans. Despite some progress, existing methods represent knowledge in the form of natural language text. Due to the long context and redundant content, they struggle to meet the robots' requirements for real-time and precise reasoning. In order to bridge this gap, we present a novel graph-based LLM, denoted as AssemMate, which consists of two stages: graph-based question answering and vision-enhanced grasp execution. The first stage enables natural language question answering on a knowledge graph, supporting human-robot interaction and assembly task planning for specific products. The second stage then utilizes the planning generated before as a target, senses stacked scenes, and executes grasping to assist with assembly. Specifically, a self-supervised Graph Convolutional Network (GCN) encodes knowledge graph entities and relations into a latent space and aligns them with LLM's representation, enabling the LLM to understand graph information. In addition, a vision-enhanced strategy is employed to address stacked scenes in grasping. Through training and evaluation, AssemMate outperforms existing methods, achieving 6.4\% higher accuracy, 3 times faster inference, and 28 times shorter context length, while demonstrating strong generalization ability on random graphs. And our approach further demonstrates superiority through robotic grasping experiments in both simulated and real-world settings. More details can be found on the project page \url{ https://github.com/cristina304/AssemMate.git}.

\end{abstract}

%%%%%%%%%%%%%%%%%%%%%%%%%%%%%%%%%%%%%%%%%%%%%%%%%%%%%%%%%%%%%%%%%%%%%%%%%%%%%%%%
\section{INTRODUCTION}

\begin{figure*}[t]
	\centering
		\includegraphics[width=2.0\columnwidth]{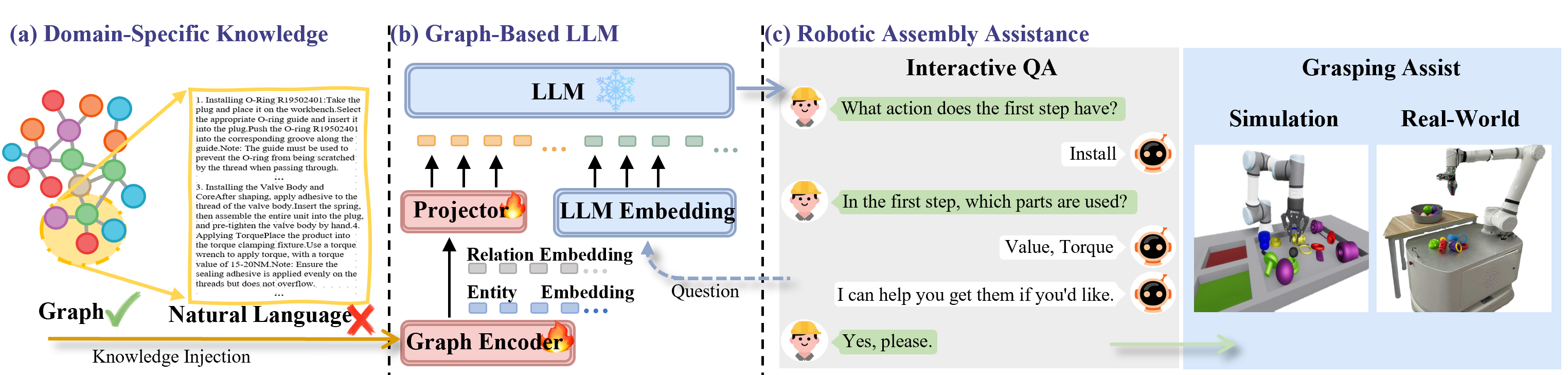}
	\caption{Our method, AssemMate, efficiently injects concise graph-structured data into LLM as external knowledge to enable interactive QA during the assembly process. It further supports grasping in stacked scenes, serving as an intelligent mate for robotic assembly assistance.}
	\label{fig:overview}
         % \vspace{-10pt}
\end{figure*}

Recently, the advancements in LLMs have shown promising results in robotic assembly assistance \cite{jiang2025prompt2act, Lim2024, Hua2025}. LLM-based robotic assembly assistance, as shown in Fig. \ref{fig:overview}, requires the injection of domain-specific knowledge into LLM, supporting real-time interaction with humans to guide accurate assembly. It also involves executing physical operations, such as grasping and installing, to improve assembly efficiency. The primary challenge lies in how to effectively and accurately inject the indispensable domain-specific knowledge into LLM. Existing methods attempt to do so by using natural language text as external knowledge. However, the long context length and redundant content of such text hinder the ability to meet the real-time and precise reasoning requirements in robotics\cite{ni2024grid}.

Compared to text, the graph is a concise and precise data structure, as shown in Fig. \ref{fig:overview}(a). There has been widespread research on transferring the text comprehension abilities of LLMs to graph structures \cite{huang2023can, li2024glbench, ren2024survey, dai2024graphpatterns}. Among them, the two most prominent approaches are representing graphs as text sequences \cite{tang2024graphgpt,chen2024llaga,ye2023language} or using Graph Neural Networks (GNNs) to encode graphs before feeding them into LLMs \cite{zhao2022learning,wang2024llms,jin2023patton}. These studies are limited to classical graph learning tasks, such as link prediction and node classification. In contrast, robotic assembly assistance requires knowledge graph question answering (KGQA) for human-robot interaction, guiding the correct assembly process and planning downstream operations. Moreover, the specific types of relationships in graphs are crucial, yet their representations in existing graph-based LLMs are limited to the simple existence or non-existence of edges. These limitations hinder the application of graph in robotic assembly assistance.

In response to these limitations, we propose AssemMate, a novel graph-based LLM that leverages GCN to vectorize graphs and encode relations for knowledge injection into LLM. Specifically, as shown in Fig. \ref{fig:overview}(b), all entities and relations in the knowledge graph are encoded via a Graph Encoder to capture both structural and semantic information, and the resulting embeddings are subsequently aligned with the LLM's representation. By injecting assembly knowledge, LLM enables human-robot interaction and assembly task planning in the form of efficient and accurate natural language KGQA. Furthermore, vision enhancement is applied to sense stacked scenes, and the robot executes the grasp to assist assembly. In addition, considering the fact that domain-specific graphs lack corresponding QA data to support network training, we propose a method for automatically constructing diverse QA datasets.

In the interactive QA phase, we use accuracy to measure single-hop QA performance, while for multi-hop QA, we use normalized LCS (nLCS) to assess sequence consistency and Weighted Jaccard Similarity (wJaccard) to evaluate content accuracy. In the grasping phase, we use the optimal planning rate (OPR) to measure the grasping planning ability in stacked scenes. Experimental results demonstrate that AssemMate achieves an accuracy of 82.1\% on single-hop QA, with an average inference time of only 0.48s. For multi-hop QA, it attains 66.7\% nLCS and 65.3\% wJaccard. The comparative experiments reveal that AssemMate surpasses existing methods with 6.4\% higher accuracy, 3 times faster inference, and 28 times shorter context length on single-hop QA, as well as achieving over 20\% higher nLCS and wJaccard on multi-hop QA. When tested on random graphs, the QA performance remains stable, validating the generalization capability of our method. For grasping tasks in both simulation and real-world scenarios, the OPR reaches 71.2\% and 64.3\%, respectively.

In summary, the main contributions of our work are:

\begin{itemize}
	\item To the best of our knowledge, we are the first to inject graph-structured data as domain-specific knowledge into LLM for robotic assembly assistance.
	\item We propose AssemMate, a pioneering approach that realizes natural language KGQA to support human-robot interaction and assembly task planning, while also enabling the modeling of specific relationships for accurate assembly assistance.
    \item Extensive experiments show that our approach outperforms existing methods with 6.4\% higher accuracy, three times faster inference, and 28 times shorter context length.
\end{itemize}

\section{Related Work}

The most closely related topics are reviewed in this section, the other robotic assembly assistant techniques, such as knowledge driven assembly model \cite{xu2024reconfigurable}, assembly control language \cite{xiao2023assembly}, can be found in the given reference paper.

\subsection{Large Language Models for Graphs}

Exploring whether LLMs can transfer their text comprehension abilities to graphs has garnered significant attention \cite{dai2024graphpatterns}. Currently, two main approaches exist: The first method involves converting graph information into natural language prompts, allowing LLMs to perform tasks without additional training, as in GraphText \cite{zhao2023graphtext} and Graph-LLM \cite{chai2023graphllm}, or introducing a projection layer as in LLaGA \cite{chen2024llaga} to further leverage the text information. The second method \cite{tang2024graphgpt,wang2024llms} uses a GNN encoder to tokenize the graph before processing it with the LLM. However, research still primarily focuses on traditional graph learning tasks, including node classification, link prediction, and graph classification, without fully leveraging the natural language capabilities of LLMs to unify these tasks into graph-based question answering and solve them with a unified paradigm. Apart from this, these methods lack explicit modeling of different relations, making them unsuitable for the practical needs of the robotics domain. 

In contrast, our AssemMate requires only basic graph triples, a fundamental and interpretable graph description form. By leveraging a self-supervised GCN to encode both entities and relations, we obtain the embedding representation of the graph, which is then integrated with the LLM through a simple and flexible projector. 

\subsection{Language-Guided Grasping}

Different to deep learning based grasping pose estimation \cite{zeng2021parametricnet, xie2024parametricnet++, huang2024sd}. Large-scale models pretrained on vast amounts of internet data have demonstrated zero-shot generalization ability to unseen scenarios \cite{achiam2023gpt,touvron2023llama}. Research on using LLMs, Vision-Language Models (VLMs), and Vision-Language-Action (VLA) models for grasping tasks is gaining significant attention. Some methods \cite{jin2024reasoning,mirjalili2024lan,tang2023graspgpt} leverage the language processing capabilities of LLMs to enhance robots' understanding of human instructions. ThinkGrasp \cite{qian2024thinkgrasp} and VLG \cite{xu2023visionlanguageaction} directly utilize the vision-language alignment capabilities of Multi-modal Large Language Models (MLLMs) to enable efficient grasping. VLA \cite{kim2024openvla,zitkovich2023rt2}, as an end-to-end approach for generating grasp poses, is also a current research hotspot, but it is greatly limited by dataset issues and the sim-to-real gap. Despite significant progress, these methods could further exploit the power of pre-trained models and place more emphasis on real-world stacked scenarios. 

On the contrary, our AssemMate determines the target object based on interactive QA, and adopts vision augmentation to compensate for the limitations of MLLMs in spatial understanding and industrial parts perception. This enables it to handle stacked situations, achieving reliable object grasping for robotic assembly assistance.

\begin{figure*}[t]
	\centering
		\includegraphics[width=2.0\columnwidth]{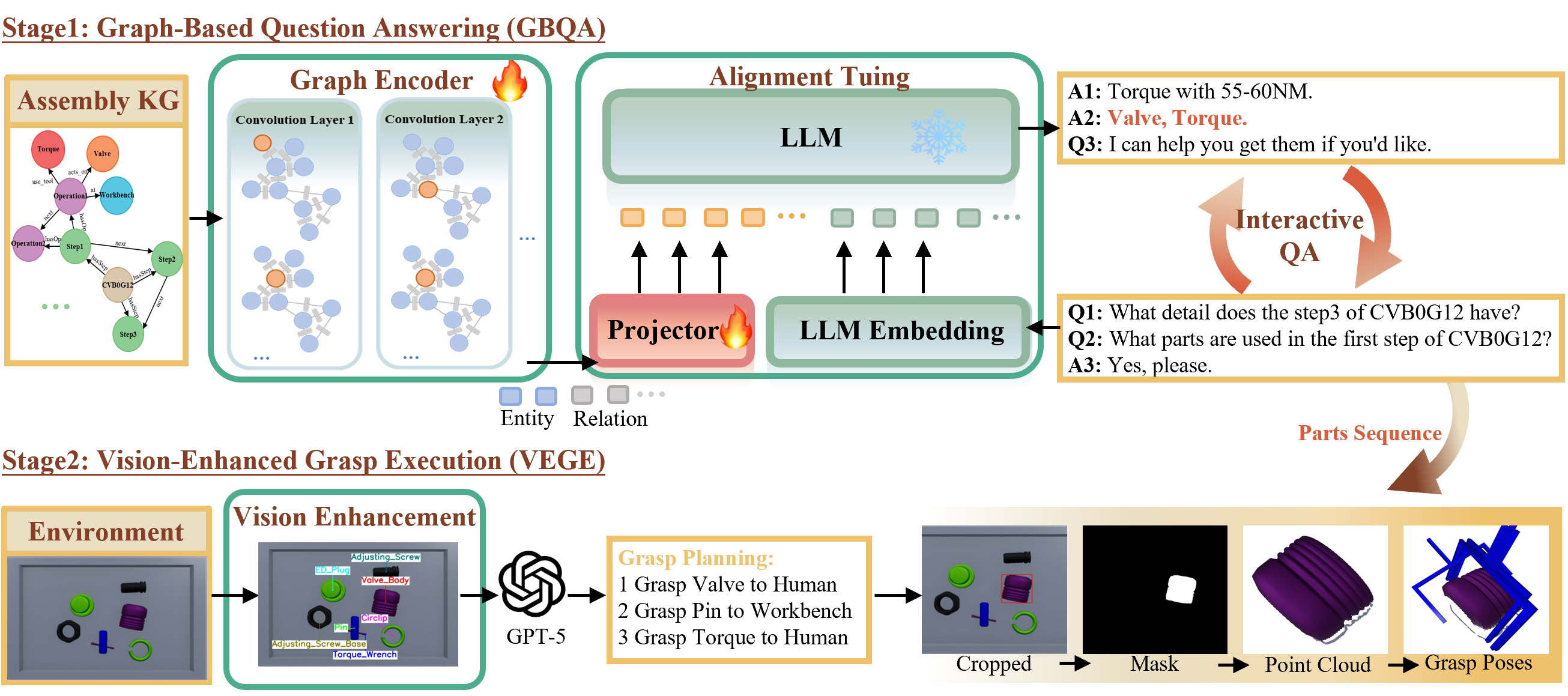}
	    \caption{Framework of AssemMate, a self-supervised GCN to encode knowledge graph entities and relations into a latent space, aligning them with LLM's representation. Based on the KGQA, VEGE leverages MLLMs with vision enhancement to sense stacked scenes and generate grasping plans, followed by segmentation and grasp poses generation.}
	\label{fig:pipeline}
         % \vspace{-10pt}
\end{figure*}

\section{PROBLEM STATEMENT}

We define the tasks that AssemMate can perform as follows: Given a human language query $Q$ and an assembly knowledge graph $G$, it can perform both single-hop and multi-hop question answering. The answers $A$ contain both assembly knowledge $K$ and the sequence of assembly parts $S$, enabling human-robot interaction and assembly task planning. Furthermore, based on the part sequence $S$, AssemMate can execute grasping operations $\hat{A}$ in stacked scenarios $Env$ to assist with human assembly. The overall process is defined as follows:

\begin{equation}
\begin{gathered}
\mathrm{AssemMate}(G, Q) \rightarrow A \\
\mathrm{If\ required\ AssemMate}(A, Env) \rightarrow \hat{A}
\end{gathered}
\end{equation}
In this process, the assembly knowledge graph $G = \langle \mathcal{N}, \mathcal{E} \rangle$, where each entity $n_i \in \mathcal{N}$ represents \emph{a part}, \emph{tool}, or \emph{workspace}. Each edge $\epsilon_i \in \mathcal{E}$ connects two entities, representing a relation type $r_i \in \mathcal{R}$, such as \emph{acts\_on}, \emph{acts\_to}, \emph{uses\_tool}, etc. The input knowledge graph can be created manually or automatically \cite{xu2025assembly}.

Additionally, the two types of question answering are defined as follows.

\begin{itemize}
	\item \textbf{Single-hop Question Answering}: Understanding the question and reaching the answer entity from a given entity through only one edge in the graph.
	\item \textbf{Multi-hop Question Answering}: Understanding the question and reaching the answer entities from a given entity by traversing multiple entities or edges in the graph.
\end{itemize}

\section{Method}

AssemMate can be divided into two stages, including Graph-Based Question Answering (GBQA) and Vision-Enhanced Grasp Execution (VEGE). The implementation details are depicted in Fig. \ref{fig:pipeline}. AssemMate, an intelligent mate designed for industrial assembly scenarios, can seamlessly understand the injected knowledge graph and interact with humans through QA to guide correct assembly. Moreover, assembly task planning can be realized within this interaction process. And AssemMate is also capable of sensing stacked scenes and executing grasping to complete the planning, achieving efficient and accurate robotic assembly assistance. 

\subsection{Graph-Based Question Answering}
Graph-Based Question Answering can be divided into two parts: Graph Encoder and Alignment Tuning. The first part encodes the information of the knowledge graph through a self-supervised learning GCN, mapping it to a latent space representation. $E_G = [e_1, e_2, \dots, e_m] \in \mathbb{R}^{m \times d'}$, where $m$ is the total number of entities $\mathcal{N}$ and relations $\mathcal{R}$ in the graph $\mathcal{G}$. The second part achieves alignment between $E_G$ and the LLM's pre-training representation by training two linear projection layers, allowing the LLM to understand the graph information.

\textbf{Graph Encoder}. Considering the complexity of relationships in industrial assembly, we use GCN on multi-relational graphs as an encoder. We adopt the graph encoding method from CompGCN \cite{vashishth2019composition}. After comparative experiments, we chose Corr \cite{nickel2016holographic} as the entity-relation composition operation, which is defined as:
\begin{equation}
r = \phi(h,t) = h \odot t,
\end{equation}
where $\phi: \mathbb{R}^d \times \mathbb{R}^d \rightarrow \mathbb{R}^d$ is a composition operator, $h \in \mathbb{R}^d$, $r \in \mathbb{R}^d$, and $t \in \mathbb{R}^d$ are the embedding vectors of head entity, relation, and tail entity, respectively. The symbol $\odot$ represents element-wise multiplication.

Additionally, we use DistMult \cite{yang2014embedding} as the score function:
\begin{equation}
\text{score}(h,r,t) = \sum_{i=1}^d h_i \cdot r_i \cdot t_i,
\end{equation}

\textbf{Alignment Tuning}. By using the GCN, the structural information of the graph is stored in a latent space, where each entity or relation has a corresponding $\mathbb{R}^d$ embedding. To validate the method's ability to learn graph structural information, we adopt an unordered knowledge graph representation method to characterize the entire graph.

Specifically, for all entities and relations in the dataset, we maintain a dictionary in random order. For a graph with a total of $m$ entities and relations, its vectorized representation is the ordered sequence of these $\mathbb{R}^d$ embeddings as they appear in the dictionary, yielding:
\begin{equation}
E_{\text{structural}} = [k_1, k_2, \dots, k_m] \in \mathbb{R}^{m \times d},
\end{equation}
where $k_i \in \mathbb{R}^d$ is the structural information vector of the $i$-th entity or relation in this graph.

Given that structural and semantic information are both important for zero-shot transfer \cite{li2024glbench}, we further encode the textual information represented by entities or relations using BERT \cite{devlin2019bert}, yielding:
\begin{equation}
E_{\text{semantic}} = [t_1, t_2, \dots, t_m] \in \mathbb{R}^{m \times d},
\end{equation}
where $t_i \in \mathbb{R}^d$ is the semantic information vector of the $i$-th entity or relation in this graph.

Then, considering that LLM alone cannot fully capture the structural and semantic information of knowledge graphs, and that discrepancies exist between the two. We propose using two separate two-layer linear projectors for Alignment Tuning:
\begin{equation}
E_{\text{structural}} = \text{softmax}(\text{MLP}_{\text{struct}}(E_{\text{structural}})),
\end{equation}

\begin{equation}
E_{\text{semantic}} = \text{softmax}(\text{MLP}_{\text{sem}}(E_{\text{semantic}})),
\end{equation}

\begin{equation}
E_G = \text{concat}(E_{\text{structural}}, E_{\text{semantic}}) \in \mathbb{R}^{m \times d'},
\end{equation}
where $d'$ is the embedding dimension of the LLM. Finally, we combine $E_G$ with the user input instruction and train the projectors on a frozen-weight LLM to enable question answering about graph knowledge.

\textbf{Loss Function}. Consistent with language autoregressive models, we use the standard cross-entropy loss function $L_{\text{ce}}$. Additionally, we incorporate a top-k sparsity penalty term $L_{\text{penalty}}$ along with a restriction mechanism:

\begin{equation}
L_{\text{penalty}} = \frac{\lambda_{\text{penalty}}}{N} \sum_{i=1}^N \sum_{j=1}^K \left( \text{logits}_{\text{top-k}}[i,j] \right)^2,
\end{equation}

\begin{equation}
L = L_{\text{ce}} + \min \left( L_{\text{penalty}},\ L_{\text{ce}} \cdot \text{max\_ratio} \right),
\end{equation}
where $\lambda_{\text{penalty}}$ controls the strength of the penalty term, $N$ is the sequence length, $K$ is the number of top-k values, $\text{logits}_{\text{top-k}}$ represents the top-k largest logits, and $\text{max\_ratio}$ limits the maximum contribution of the penalty term to the overall loss function.

\begin{table}[t]
\centering
\caption{Example Questions for Single-hop QA}
\begin{tabular}{@{}ll@{}}
\toprule
\textbf{Category} & \textbf{Example Questions} \\ \midrule
\textbf{Action}   & What action is performed in step1 of the CV01S? \\
                  & What action is done by step1 of the CV01S ? \\
                  & What action does step1 of the CV01S carry out? \\ \midrule
\textbf{Tools}    & What tool is used in step1 of the CV01S ? \\
                  & Which tool is utilized by step1 of the CV01S ? \\
                  & Can you name the tools used by step1 of the CV01S ? \\ \midrule
\textbf{Workspace} & To what position does step1 of the CV01S act? \\
                   & Where does step1 of the CV01S act to? \\
                   & To which location does step1 of the CV01S act? \\ \midrule
\textbf{Detail}   & What detail does step1 of the CV01S provide? \\
                  & What detail is given for step1 of the CV01S? \\
                  & What detail is associated with step1 of the CV01S ? \\ \bottomrule
\end{tabular}
\label{tab:qa}
\end{table}

\begin{table*}[!t]
\centering
\caption{Comparison experiment against the use of natural language text as domain-specific knowledge injection.}
\begin{tabular}{@{}c c c c c c@{}}
\toprule
\multirow{2}{*}{\parbox[c][2em][c]{3cm}{\centering Method}} & \multicolumn{3}{c}{Single-hop QA} & \multicolumn{2}{c}{Multi-hop QA} \\ 
\cmidrule(lr){2-4} \cmidrule(lr){5-6}
       & In-context Length ↓ &  Infer Time (s) ↓ &Acc. (\%) ↑  & nLCS (\%) ↑ & wJaccard (\%) ↑ \\ 
\midrule
Text Triple \cite{kim2023kggpt}             & 3510 & 1.50 & 75.7  & 45.7 & 43.4 \\
Text Triple + Fine-tuning \cite{feng2025grip}        & 3510 & 1.51 & 78.4  & 47.0 & 45.2 \\
Document \cite{hua2025integration}        & 764  & 0.85 & 50.0  & 46.3 & 43.4 \\
Document + Fine-tuning \cite{su2019generalizing,basem2024optimized} & 764  & 0.83 & 75.8  & 53.6 & 53.8 \\
\textbf{Ours (Graph + Projector)} & \textbf{125}  & \textbf{0.48} & \textbf{82.1} & \textbf{66.7} & \textbf{65.3} \\
\bottomrule
\end{tabular}
\label{tab:comparison}
\end{table*}

\subsection{Vision-Enhanced Grasp Execution}

In industrial assembly, grasping in cluttered scenes is a critical issue that must be addressed. To tackle this, we apply vision enhancement, which leverages the powerful reasoning capabilities of MLLMs to analyze the information in the current scene and generate optimal grasping plans. After that, VEGE detects the position of the target objects in the scene and generates the optimal grasp pose.  

\textbf{Vision Enhancement}. Aiming to improve the spatial perception and cross-modal alignment abilities of MLLMs, we introduce a text- and arrow-based visual prompt enhancement method, which annotates object names with text labels and uses arrows to connect the text labels to the object's center, as shown in Fig. \ref{fig:real}. This improves the cross-modal alignment capability of the model. To reduce occlusion of the original image, text labels are placed in background areas that are weakly related to the task, and the scanning circle strategy is employed to select the most suitable position for placing the text labels.

\section{Experiments}
\subsection{Dataset}
We propose an automated method for constructing diverse QA datasets. Specifically, for a given knowledge graph, we decompose it entirely into \(<head \ entity, \ relation, \ tail \ entity>\) triples. Then, we use chain-of-thought (CoT) reasoning \cite{xuan2024pink} and few-shot \cite{Brown2020} strategies to guide LLM in generating rich and diverse single-hop and multi-hop QA pairs based on these triples, involving \emph{actions}, \emph{tools}, \emph{workspaces}, \emph{subassemblies}, \emph{assemblies}, \emph{object attributes}, and \emph{assembly details} during the assembly process. 

Examples of single-hop QA are shown in Table~\ref{tab:qa}, while examples of multi-hop QA are as follows:

\begin{itemize}
    \item \emph{List the parts and tools used in step1 of the CV01S, in the order they are used.}
    \item \emph{In step1 of the CV01S, what are the parts and tools involved sequentially?}
    \item \emph{Identify the sequential parts and tools used in step1 of the CV01S.}
\end{itemize}
where \emph{CV01S} represents the type of the pressure-reducing valve. The answers for multi-hop QA consist of sequences of objects involved in the assembly in order, which serve as plans for downstream grasping operations.

Assembly knowledge graphs are obtained from multiple information sources, 
such as assembly process documents and 3D models of assemblies. We used the 
assembly graphs of 12 pressure-reducing valves as the training set, applying a strategy that randomly splits the entire graph into subgraphs, thereby expanding the training dataset to 148 graphs to improve generalization. Using our QA construction strategy, we generated a dataset of 22,000 examples for training. The testing set consists of the assembly graphs of 6 products, where the entire product graphs are used as input to align with real-world application scenarios.

In the Alignment Tuning stage, both the question and the corresponding graph knowledge are simultaneously injected into the LLM as input. To better distinguish the information, we adopt the following instruction format:
\emph{\textless kg\_start\_token\textgreater \{graph information\} \textless kg\_end\_token\textgreater} \\
\emph{\# Task: Based on the assembly knowledge graph information above, answer the question} \\
\emph{\# Question \{question\}} \\
Here, \emph{graph information} refers to the graph embedding \(E_G\). The other parts of the instruction are encoded by the LLM's embedding layer and concatenated with \(E_G\).

\subsection{Training details}

For the Graph Encoder, we employed CompGCN \cite{vashishth2019composition}, with the hyperparameters set as follows: score\_func = distmult, opn = corr, init\_dim = 128, embed\_dim = 768, and gcn\_layer = 2. In the Alignment Tuning stage, we used LLaMA3-8b-4bit \cite{touvron2023llama} as the base model. The implementation is carried out using the PyTorch framework with the AdamW optimizer, an initial learning rate of $1\times10^{-4}$, and cosine decay scheduling. The projector layer consists of two 2-layer Feed-Forward Networks (FFNs). The model is trained for a maximum of 100 epochs with a batch size of 1, across three NVIDIA RTX 3090 GPUs. The loss function parameters are set as 
$\lambda_{\text{penalty}}=1\times10^{-5}$, 
$\text{top\_k}=10$, 
and $\text{max\_ratio}=0.1$.

\begin{table}[t]
\centering
\caption{ablation study on AssemMate module. w/o, without}
\setlength{\tabcolsep}{3pt}
\begin{tabular}{@{}c c c c c@{}}
\toprule
\multirow{2}{*}{\centering Method} & \multicolumn{2}{c}{Single-hop QA} & \multicolumn{2}{c}{Multi-hop QA} \\ 
\cmidrule(lr){2-3} \cmidrule(lr){4-5}
       & Infer Time (s) ↓ & Acc. (\%) ↑ & nLCS (\%) ↑ & wJaccard (\%) ↑ \\ 
\midrule
w/o GCN                    & 0.48 & 78.9 & 56.0 & 53.6 \\
w/o relation      & 0.64  & 18.4 & 1.8 & 1.3 \\
w/o entity        & 0.65  & 21.1 & —    & —    \\
\textbf{Ours}             & \textbf{0.48} & \textbf{82.1} & \textbf{66.7} & \textbf{65.3} \\
\bottomrule
\end{tabular}
\label{tab:ablation}
\end{table}

\subsection{Evaluation}
\textbf{Metrics}.
For single-hop QA, evaluation is based on answer accuracy (Acc). 
For multi-hop QA, we use normalized LCS (nLCS) to measure the consistency of sequence order 
and Weighted Jaccard Similarity (wJaccard) to assess content consistency. For the grasping experiments in stacked scenarios, we adopt two metrics: 
average step and optimal planning rate (OPR), defined as
\begin{equation}
\text{OPR} = \frac{\text{Number of Optimal Plans}}{\text{Number of Actual Operations}}
\end{equation}

\textbf{Comparative Experiments}. As previously mentioned, existing methods often represent domain-specific knowledge in the form of natural language text. To ensure the credibility of our comparative experiments, we compare our approach with existing methods that perform QA tasks based on external knowledge. These methods can be categorized into four types:
\begin{itemize}
    \item \textbf{Document:} Using plain textual documents directly as prompts for LLMs \cite{hua2025integration}.
    \item \textbf{Document + Fine-tuning:} Fine-tuning LLMs on the basis of these textual documents \cite{su2019generalizing,basem2024optimized}.
    \item \textbf{Text Triple:} Converting graph-structured data into natural language text as input to LLMs \cite{kim2023kggpt}.
    \item \textbf{Text Triple + Fine-tuning:} Fine-tuning LLMs on graph-derived textual representations \cite{feng2025grip}.
\end{itemize}
In these experiments, we consistently adopt LoRA \cite{hu2022lora} as the method for fine-tuning.

As shown in Table~\ref{tab:comparison}, compared to existing methods, our approach demonstrates clear advantages in both efficiency and accuracy. This can be intuitively attributed to the fact that, by encoding the graph knowledge via GCN into embeddings, the required context length is reduced by 28 times, and inference is accelerated by 3 times with only 0.48s, which greatly reduces computational resource requirements, facilitates edge deployment, and better suits the needs of real-world robotic assembly assistance. Moreover, by training only the projector layer, our method achieves an accuracy of 82.1\% on single-hop QA, 66.7\% nLCS, and 65.3\% wJaccard on multi-hop QA. It surpasses existing methods with 6.4\% higher accuracy on single-hop QA, and 21\% higher nLCS and 21.9\% higher wJaccard on multi-hop QA.

In addition, in the comparative experiments, we observed that assembly process documents contain a large amount of distracting information, such as symbols and referential expressions, which results in the lowest quality of injected knowledge when the LLM is not fine-tuned. In contrast, although triples, like graphs, are structured data and conducive to querying, they suffer from significant entity redundancy. Specifically, the same entity may be connected by multiple edges, causing it to be repeatedly represented in the triples, which increases the context length and consequently prolongs the inference time.

\begin{table}[t]
\centering
\caption{Testing generalization ability on random graphs.}
\begin{tabular}{@{}c c c c@{}}
\toprule
\multirow{2}{*}{\centering Dataset} & Single-hop QA & \multicolumn{2}{c}{Multi-hop QA} \\ 
\cmidrule(lr){2-2} \cmidrule(lr){3-4}
        & Acc. (\%) ↑ & nLCS (\%) ↑ & wJaccard (\%) ↑ \\ 
\midrule
Testing Graph        & 82.1 & 66.7 & 65.3 \\
Random Graph    & 75.9 & 56.9 & 55.6 \\
\bottomrule
\end{tabular}
\label{tab:random}
\end{table}

\begin{figure*}[t]
	\centering
		\includegraphics[width=2.0\columnwidth]{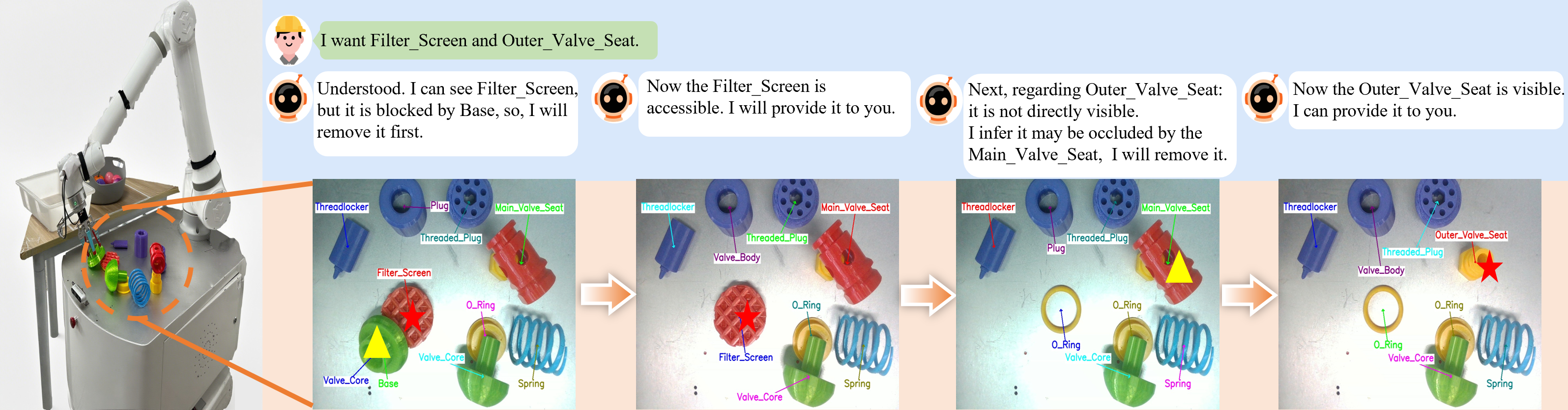}
	    \caption{In real-world experiments, AssemMate dynamically determines the optimal object to grasp based on the current scene, where the yellow triangle indicates the current considered object and the red star marks the target object.}
	\label{fig:real}
         % \vspace{-10pt}
\end{figure*}
\textbf{Ablation Study}. As mentioned earlier, our graph knowledge $E_G$ consists of two components: $E_{\text{structural}}$ and $E_{\text{semantic}}$. 
The ablation of the GCN module essentially removes the $E_{\text{structural}}$ information. 
As shown in Table~\ref{tab:ablation}, this leads to a performance drop in both single-hop and multi-hop QA tasks, 
particularly in multi-hop QA, where the decrease is nearly 10\%, compared to around 3\% in single-hop QA. 
This aligns with the fact that multi-hop reasoning relies more heavily on structured information than single-hop reasoning, 
highlighting the necessity of the GCN module. 
In addition, as indicated in Table~\ref{tab:ablation}, removing the relation embedding results in catastrophic performance degradation, 
underscoring the critical importance of encoding specific relations in robotic applications.

\textbf{Generalization to Random Graphs}. To validate the generalization of our knowledge injection method, we construct random graphs at both the step and operation levels for testing. And the results in Table~\ref{tab:random} demonstrate that our method performs excellently on these tests, with only a minor decrease of 8.5\% in single-hop QA accuracy, and 10.8\% and 14.3\% in nLCS and wJaccard for multi-hop QA, respectively, compared to the testing graph. This indicates that our approach conducts reasoning over the injected graph embedding rather than relying solely on the memorization capability of LLM, suggesting promising potential for extending our method to broader domains.

\begin{figure}[t]
	\centering
		\includegraphics[width=1.0\columnwidth]{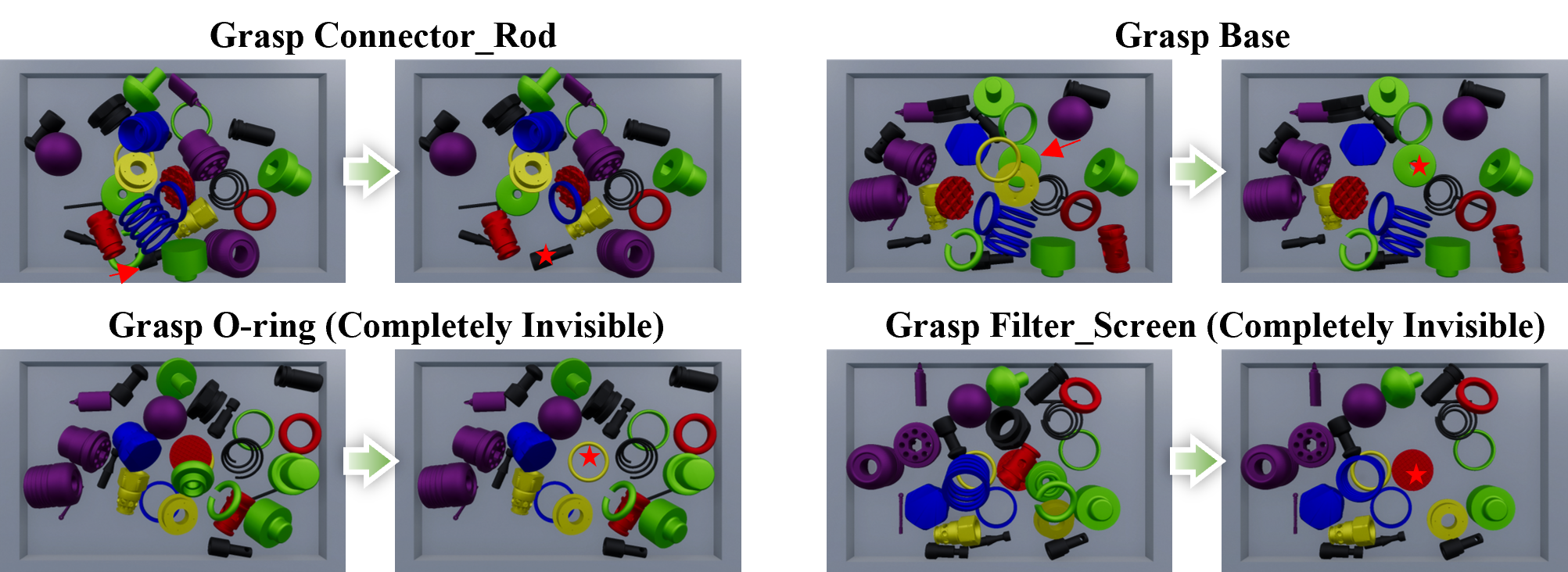}
	    \caption{The heavily stacked scenarios within the simulation environment, the target object is marked with a red star.}
	\label{fig:sim}
         % \vspace{-10pt}
\end{figure}
\subsection{Simulation and Real-World Experiment}

\textbf{System Setup}. The VEGE pipeline consists of a vision-enhanced reasoning module, an object perception module, and a grasp synthesis module. Vision-enhanced reasoning is performed using the GPT-5 API~\cite{openai2025gpt5}. The model takes the task description together with the enhanced-scene image generated by the vision enhancement method described in Section~IV-B as input, and outputs an optimal grasping plan. Given the selected target from the plan, object-level perception is carried out using YOLOv11~\cite{khanam2024yolov11} for 2D object detection, followed by SAM2~\cite{ravi2024sam2} for precise instance segmentation. The resulting segmentation mask, together with depth information, is provided as input to GraspNet-1Billion~\cite{fang2020graspnet} to generate candidate grasp poses on the target object. 
The grasp with the highest predicted success score is selected and executed by the robot.
To improve robustness under cluttered conditions, we collected 2,000 stacked-scene images in a simulation environment to fine-tune the YOLOv11 detector for 2D localization. 

Our simulation environment is built on the Unity engine, using a UR3 robotic arm equipped with a Robotiq 2f-85 two-finger gripper. For real-world experiments, we employ an ELITE ROBOTS EC612 Collaborative Robot with a WHEELTEC four-finger flexible electric gripper and an Intel RealSense D435i depth camera.

We respectively evaluate grasping performance under the answers of single-hop and multi-hop QA as upstream planning. For each QA, 10 planning cases are tested, executed 10 times in both simulation and real-world setups. The stacked scenes consist of 22 randomly stacked objects, which are kept consistent between simulation and real-world experiments.

In the real-world grasping experiments, we 3D-printed 22 components and tools based on the 3D model of the pressure-reducing valve to serve as graspable objects. 

\begin{table}[t]
\centering
\caption{Evaluation of grasping performance in stacked scenarios.}
\resizebox{\linewidth}{!}{
\begin{tabular}{lcccc}
\toprule
 & \multicolumn{2}{c}{Single-hop} & \multicolumn{2}{c}{Multi-hop} \\
\cmidrule(lr){2-3} \cmidrule(lr){4-5}
 & Simulation & Real & Simulation & Real \\
\midrule
OPR (\%) $\uparrow$ & 71.2 & 64.3 & 64.0 & 52.4 \\
Average Step $\downarrow$ & 3.1 & 3.9 & 6.2 & 7.3 \\
\bottomrule
\end{tabular}
}
\label{tab:grasping_results}
\end{table}
\textbf{Results and Analyze}. The real-world grasping experiment is shown in Fig. \ref{fig:real}. AssemMate dynamically determines the optimal object to grasp based on the current stacked scene, enabling it to sense scenarios where objects are stacked or even fully occluded. This shows that AssemMate can remove obstructing objects and grasp the user-desired one in a stacked industrial environment, assisting in assembly. The results of the grasping experiments are presented in Table~\ref{tab:grasping_results}. Under upstream planning with single-hop QA, the optimal planning rate for grasping in simulation reaches 71.2\%, while for multi-hop QA planning of object sequences, it reaches 64.0\%. This demonstrates the effectiveness and efficiency of our approach in sensing stacked scenes and executing grasping. In real-world experiments, due to object collisions during the grasping process and lighting effects during detection, the performance is lower than in simulation by 6.9\% for single-hop QA and 11.6\% for multi-hop QA, showing good sim-to-real transfer capability. Complex stacked scenarios in simulation are illustrated in Fig.\ref{fig:sim}, where AssemMate efficiently and accurately removes occluding objects surrounding the target.

\section{CONCLUSIONS}
In this paper, we introduce a reliable mate, AssemMate, which, for the first time in the field of robotic assembly assistance, leverages graph\textemdash the concise and precise data structure\textemdash as the means to represent assembly knowledge and efficiently inject it into LLM. This enables LLM to acquire knowledge absent during pretraining and respond to human queries to support interactive QA for correct assembly guidance. Moreover, we employ vision enhancement techniques to fully exploit the capabilities of MLLMs, allowing execution of grasping in stacked scenarios, thereby improving assembly efficiency. Experimental results demonstrate that our method outperforms existing approaches on the same tasks in terms of accuracy, efficiency, and resource consumption, and can be easily extended to private graphs and specific tasks across various domains.

\section{acknowledgement}
This work is supported by National Natural Science Foundation of China (Grant No. 92467204).
%\addtolength{\textheight}{-12cm}   % This command serves to balance the column lengths
                                  % on the last page of the document manually. It shortens
                                  % the textheight of the last page by a suitable amount.
                                  % This command does not take effect until the next page
                                  % so it should come on the page before the last. Make
                                  % sure that you do not shorten the textheight too much.

%%%%%%%%%%%%%%%%%%%%%%%%%%%%%%%%%%%%%%%%%%%%%%%%%%%%%%%%%%%%%%%%%%%%%%%%%%%%%%%%

%%%%%%%%%%%%%%%%%%%%%%%%%%%%%%%%%%%%%%%%%%%%%%%%%%%%%%%%%%%%%%%%%%%%%%%%%%%%%%%%

%%%%%%%%%%%%%%%%%%%%%%%%%%%%%%%%%%%%%%%%%%%%%%%%%%%%%%%%%%%%%%%%%%%%%%%%%%%%%%%%

\bibliographystyle{IEEEtran}
\bibliography{root}
\end{document}